\documentclass[10pt,twocolumn,letterpaper]{article}

\usepackage{cvpr}
\usepackage{graphicx} 
\usepackage{multirow}
\usepackage{adjustbox}
\usepackage{bbm}
\usepackage{bibentry}
\usepackage{booktabs}
\usepackage{amsmath}
\usepackage{color}

\definecolor{cvprblue}{rgb}{0.21,0.49,0.74}
\usepackage[pagebackref,breaklinks,colorlinks,allcolors=cvprblue]{hyperref}

\def\paperID{907} 
\def\confName{CVPR}
\def\confYear{2025}

\title{\large SeMi: When Imbalanced Semi-Supervised Learning Meets Mining Hard Examples}

\author{
Yin Wang$^{1}$, Zixuan Wang$^{2}$, Hao Lu$^{2}$, Zhen Qin$^{1}$, Hailiang Zhao$^{1}$, Guanjie Cheng$^{1}$, \\
Ge Su$^{1}$, Li Kuang$^{3}$, Mengchu Zhou$^{4}$, Shuiguang Deng$^{1}$ \\
$^{1}$Zhejiang University, $^{2}$The Hong Kong University of Science and Technology, \\
$^{3}$Central South University, $^{4}$Zhejiang Gongshang University \\
{\tt\small waynewang@zju.edu.cn, zwanggk@connect.ust.hk, haolu@connect.hkust-gz.edu.cn,} \\
{\tt\small zhenqin@zju.edu.cn, hliangzhao@zju.edu.cn, chengguanjie@zju.edu.cn,} \\
{\tt\small suge@zju.edu.cn, kuangli@csu.edu.cn, mengchu@gmail.com, dengsg@zju.edu.cn}
}

\begin{document}
\maketitle
\begin{abstract}
Semi-Supervised Learning (SSL) can leverage abundant unlabeled data to boost model performance. However, the class-imbalanced data distribution in real-world scenarios poses great challenges to SSL, resulting in performance degradation. Existing class-imbalanced semi-supervised learning (CISSL) methods mainly focus on rebalancing datasets but ignore the potential of using hard examples to enhance performance, making it difficult to fully harness the power of unlabeled data even with sophisticated algorithms. To address this issue, we propose a method that enhances the performance of Imbalanced \underline{\textbf{Se}}mi-Supervised Learning by \underline{\textbf{Mi}}ning Hard Examples (SeMi). This method distinguishes the entropy differences among logits of hard and easy examples, thereby identifying hard examples and increasing the utility of unlabeled data, better addressing the imbalance problem in CISSL. In addition, we maintain a class-balanced memory bank with confidence decay for storing high-confidence embeddings to enhance the pseudo-labels' reliability. Although our method is simple, it is effective and seamlessly integrates with existing approaches. We perform comprehensive experiments on standard CISSL benchmarks and experimentally demonstrate that our proposed SeMi outperforms existing state-of-the-art methods on multiple benchmarks, especially in reversed scenarios, where our best result shows approximately a 54.8\% improvement over the baseline methods.
\end{abstract}    
\section{Introduction}
\label{sec:intro}


Semi-supervised learning (SSL) improves model performance using unlabeled data, but its assumption of uniform data distribution often doesn't align with reality. In practice, unevenly distributed data can bias predictions, favoring majority classes and underperforming on minority ones.


\begin{figure}[ht]
  \centering
  \includegraphics[width=\linewidth]{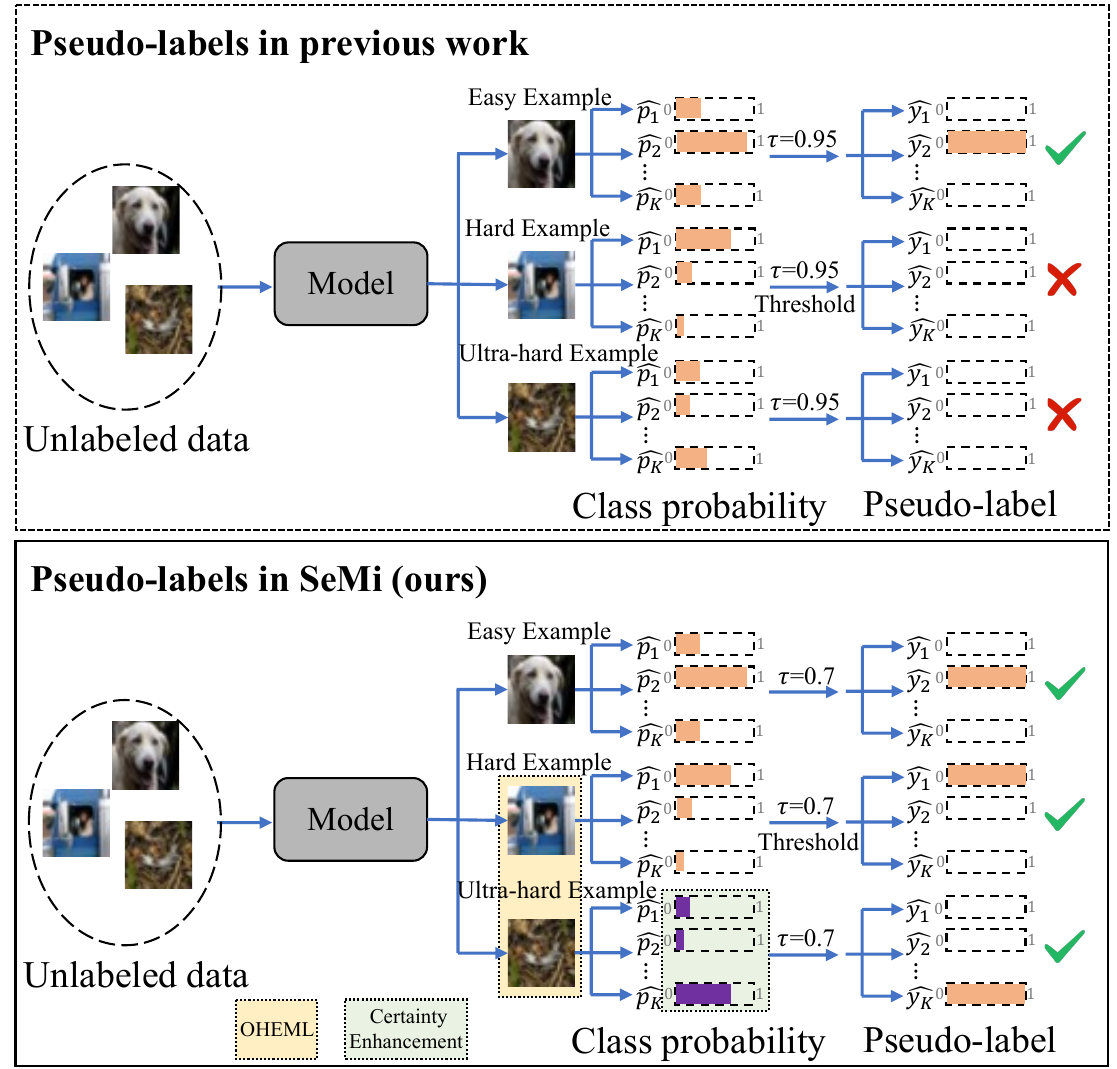}
  \caption{Differences between previous CISSL methods and our method (SeMi) in generating pseudo-labels. The previous method used a high threshold, missing opportunities to learn from hard examples, possibly from the \textit{tail}. Our method effectively uses hard examples with techniques like online hard examples mining and learning (OHEML) and pseudo-label certainty enhancement.}
  \label{fig：glimpse}
\end{figure}

Recently, more studies focus on Class Imbalanced Semi-Supervised Learning (CISSL), and propose such methods as re-sampling~\cite{wei2021crest,cao2019learning,kang2019decoupling,zhou2020bbn, wang2022imbalanced}, re-weighting~\cite{peng2023dynamic, ren2018learning, cui2019class, alshammari2022long, chen2023area, lai2022smoothed}, logits adjustment~\cite{menon2020long, wei2023towards, cao2019learning}, contrastive learning~\cite{zhou2024continuous,kang2020exploring, cui2021parametric, zhu2022balanced, du2024probabilistic} and decoupled learning~\cite{fan2022cossl} ones. These methods alleviate the problem of prediction bias for minority class samples to some extent. However, the value of unlabeled data for improving models' performance has not been fully exploited. We refer to the hard-to-learn examples in each class and the examples in minority classes as hard examples, which are crucial for improving the model's generalization and discriminative abilities. As depicted in Figure~\ref{fig：mask_prob}, we record the masked probability and used accuracy of pseudo-labels at the epoch with the best accuracy. The masked probability indicates the likelihood of assigning pseudo-labels to unlabeled data, while the used accuracy reflects the consistency of pseudo-labels with ground truth. We find that the masked probability is typically below 80\%, meaning many samples remain unused. As shown in Figure~\ref{fig：glimpse}, in CISSL, existing methods typically set a high confidence threshold (e.g., 0.95) when generating pseudo-labels, which suppresses the learning of hard examples with confidence below this threshold. However, these hard examples are often from tail classes, making them critical for improving performance on tail categories. Intuitively, lowering the confidence threshold can enable access to more samples, but directly utilizing the pseudo-labels with low confidence can seriously jeopardize the model's discriminative ability.

This paper explores an essential but under-attended CISSL problem: how to mine and utilize hard examples to improve model performance. Our study is based on three main findings: i) the samples in the minority class are hard examples, and a high confidence threshold for pseudo-labels tends to make it impossible to utilize them to improve the model performance. For the minority classes with less labeled data, mining more such samples from unlabeled data is crucial to improve the model performance. ii) Lowering the confidence threshold for pseudo-labels allows the existing methods to access more samples but increases their uncertainty and reduces their robustness. iii) Introducing a class-balanced memory bank can enhance the certainty and accuracy of pseudo-labels based on embeddings that yield robust semantic predictions. However, if there is no limitation on the embeddings to be pushed into the memory bank, embeddings with low confidence are also inserted into the memory bank. These embeddings are likely to be from the minority classes, and when the class-balanced classifier learns these embeddings, they further impair the certainty of the pseudo-labels for the minority classes. 

To address the above three findings, we propose a novel FixMatch-based method termed as SeMi, which stands for Semi-Supervised Learning by Mining Hard Examples, with three ideas. i) In order to access more hard examples, we moderately lower the confidence threshold. Samples above the threshold are reweighted to guide the model in focusing on the hard examples and improving the discriminative ability of the minority classes. In addition, for those ultra-hard samples that are still below the threshold, we accelerate the model to learn them by aligning the strong and weak views. ii) To compensate for the certainty of the pseudo-labels after lowering the confidence threshold, SeMi employs the dynamic mixing of a) semantic pseudo-labels generated by the embedding prototype and b) pseudo-labels generated by the classifier, which helps to improve the robustness and accuracy of the mixed pseudo-labels. iii) In contrast to the standard class-balanced memory bank, our memory bank inserts embeddings with high-confidence blocks and low-confidence embeddings, as the latter have high-uncertainty pseudo-labels that can harm the model's discriminative ability. At a higher level, our first two ideas leverage unlabeled data to support the model in learning more generalizable representations. Our last idea is used to further exploit these high-quality representations and thus generate more accurate pseudo-labels. The simple yet effective approach performs well in several CISSL benchmarks.

This work intends to make the following novel contributions to the field of CISSL:
\begin{enumerate}[1)]
    \item To the best of our knowledge, this is the first work to study how to leverage hard examples to enhance model generalization in imbalanced semi-supervised learning.
    \item  We propose SeMi that leverages unlabeled data to mine hard samples to enhance the representation of minority classes and exploits high-confidence embeddings to enhance the robustness and accuracy of pseudo-labels.
    \item We perform extensive experiments on standard CISSL benchmarks, demonstrating the state-of-the-art results obtained by SeMi on multiple benchmarks.
\end{enumerate}
\section{Related Work}
\label{sec:formatting}

\textbf{Semi-supervised learning.} Semi-supervised learning aims to train models using a few labeled and rich unlabeled data~\cite{berthelot2019remixmatch, sohn2020fixmatch, zhang2021flexmatch}. SSL effectively leverages unlabeled data, reducing data labeling costs and improving the model's generalization ability. In recent years, many methods have demonstrated effectiveness, such as mean teacher~\cite{tarvainen2017mean,tang2021humble,xu2021end}, pseudo labeling~\cite{arazo2020pseudo,chen2023softmatch,rizve2021defense,yan2023dml, li2024learning}, threshold adjustment~\cite{guo2022class, wang2022freematch, yu2023inpl, li2023instant, rizve2021defense} and consistency regularization~\cite{fan2023revisiting,kuo2020featmatch,xie2020unsupervised}. Several notable methods, including ReMixMatch~\cite{berthelot2019remixmatch}, FlexMatch~\cite{zhang2021flexmatch}, and FixMatch~\cite{sohn2020fixmatch}, have advanced semi-supervised learning by leveraging label distribution alignment, dynamic thresholding for learning difficult samples, and consistency regularization with pseudo-labeling to enhance model robustness. However, SSL assumes that labeled data and unlabeled data are uniformly distributed. In realistic scenarios, imbalanced labeled data and unlabeled data skew the model predictions heavily towards the majority class, leading to deterioration in performance for the minority class.

\noindent\textbf{Imbalanced semi-supervised learning.} Recent studies show great promise concerning solving class imbalanced semi-supervised learning. For instance, ABC~\cite{lee2021abc} uses an auxiliary balanced classifier and decouples the representations from the classifier by combining Bernoulli distribution for sampling majority and minority class samples. Crest~\cite{wei2021crest} improves precision for the minority class by sampling more credible data predicted to belong to the minority class, expanding the dataset. DARP~\cite{kim2020distribution} instead proposes to reﬁne pseudo-labels via convex optimization on labeled distribution. Cossl~\cite{fan2022cossl} developed a less biased classifier by enhancing the diversity of features in tail samples using the TFE technique. DASO~\cite{oh2022daso} employs a blending of semantic pseudo-labeling and linear labeling to alleviate the linear classifier bias towards the majority class. BMB~\cite{peng2023bmb} improves feature balance by constructing a memory bank for the auxiliary classifier, which learns more balanced features. ACR~\cite{wei2023towards} introduces dynamic logit adjustment to enhance performance on minority classes when the unlabeled label distribution is unknown. BEM~\cite{zheng2024bem} combines class-balanced data mixing and entropy-based strategies to tackle class imbalance and uncertainty in long-tailed semi-supervised learning. Existing studies ignore mining more minority-class samples from unlabeled data to improve the model's predictive performance on minority classes.

\begin{figure*}[ht]
  \centering
  \includegraphics[width=0.98\linewidth]{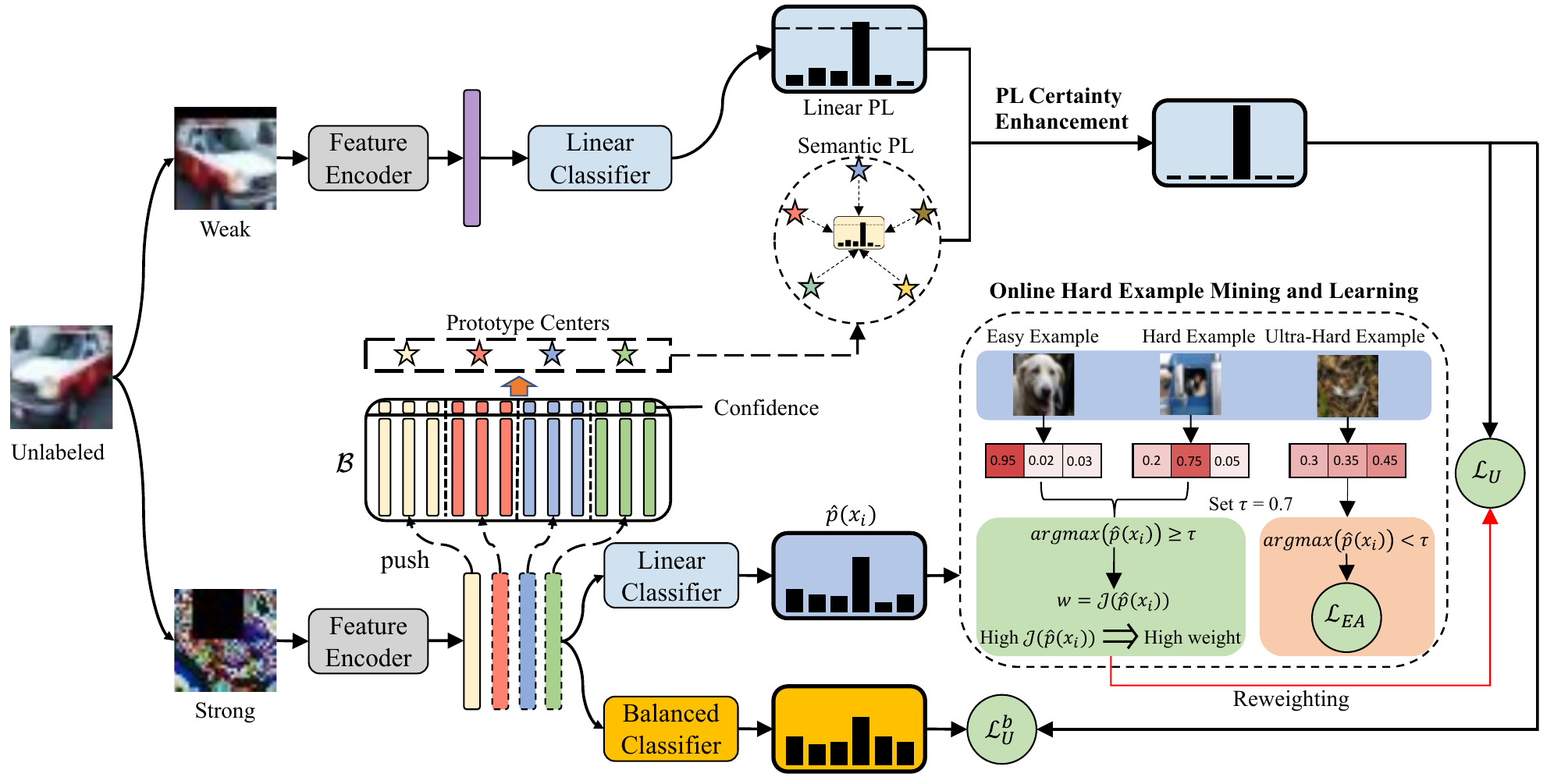}
  \caption{The pipeline of the SeMi framework. The unlabeled data is transformed into weak and strong views by image augmentation. In the strong views branch, the generated features are pushed into the cells only when their confidences reach the Balanced Confidence Decay Memory Bank's ($\mathcal{B}$) threshold. Then, the prototype centers for each category can be calculated, and semantic pseudo-labels for query embeddings are obtained. Meanwhile, the linear pseudo-labels generated from the weak views branch are mixed with the semantic pseudo-labels to get the certainty-enhanced pseudo-labels. Online hard example mining and learning to discriminate the hardness of strong views and give larger weights to the hard examples. For ultra-hard samples, the Embedding Align approach can accelerate the learning. In addition, a balanced classifier is maintained to yield predictions that are more friendly to the tail classes.}
  \label{fig:semi}
\end{figure*}
\section{Proposed Method}
In this paper, we propose the SeMi framework, shown in Figure~\ref{fig:semi}. It improves learning from hard samples and enhances pseudo-label reliability. The Online Hard Example Mining and Learning (OHEML) algorithm efficiently utilizes challenging samples. The Balanced Confidence Decay Memory Bank increases the probability of embedding high-certainty features. The Pseudo-Label Certainty Enhancement (PLCE) algorithm enhances pseudo-label robustness and accuracy by integrating semantic feature centers.
\subsection{Preliminaries}
\textbf{Problem setup.} In CISSL, suppose we have a labeled dataset $X^{(l)}=\{(x^{(l)}_{i}, y^{(l)}_{i})\}$ of size $N$ and an unlabeled dataset $X^{(u)}=\{(x^{(u)}_{j})\}$ of size $M$, where $x^{(l)}_i$, $x^{(u)}_j \in \mathbb{R}^d$ are training data, $y^{(l)}_{i} \in \{0, 1\}^K$ is ground truth for labeled data, and $K$ is the size of label space. In the CISSL setting, a small amount of data is labeled and a large amount of data is unlabeled, and they are both unbalanced. Let $N_k$ and $M_k$ denote the numbers of labeled and unlabeled examples in class k, respectively. Let  ${\textstyle \sum_{k=1}^{K}} N_k = N$ and ${\textstyle \sum_{k=1}^{M}} M_k = M$. The degree of imbalance is also critical. We assume that the $K$ classes are sorted in descending order, i.e. $N_1>N_2>\cdots>N_K$. The imbalanced ratio denoted as $\gamma _l=\frac{N_1}{N_K}$ for labeled data. In realistic scenarios, we do not have access to the prior distribution of unlabeled data, so the imbalance ratio of unlabeled data is represented by the estimated labels as $\gamma _u=\frac{{max_k}M_k}{{min_k}M_k}$. Our goal of CISSL is to utilize $X^{(l)}$ and $X^{(u)}$ to train a standard classifier $f:\mathbb{R}^d \to \{0,1\}^K$ and a class-balanced classifier $\widetilde{f}:\mathbb{R}^d \to \{0,1\}^K$ parameterized by $\theta$. Detailed symbol definitions are in Appendix {\color{red}G}.

Some typical SSL frameworks, e.g., Fixmatch, FlexMatch, and ReMixMatch, are the basis of the CISSL framework. Take FixMatch as an example here, with labeled and unlabeled base objective functions:
\begin{equation}
\mathcal{L}_S= \frac{1}{B_l}\sum_{i=1}^{B_l}  \mathcal{H} (f(x^{(l)}_i;\theta),y^{(l)}_i)
\end{equation}
\begin{equation}
\mathcal{L}_U= \frac{1}{B_u}\sum_{j=1}^{B_u}  \mathcal{M}(x^{(u)}_j) \mathcal{H} (f(\mathcal{A}_s( x^{(u)}_j);\theta),q_j)
\end{equation}
where $ \mathcal{H}$ is the cross-entropy loss. $B_l$, $B_u$ denote the batch size of labeled and unlabeled data. The mask function is represented as $\mathcal{M}(x^{(u)}_j):=\mathbbm{1}(max(\delta(f(x^{(u)}_j)))>\tau)$. $\mathbbm{1}(\cdot)$ denotes indicator function. $\delta$ denotes the softmax function. $\tau$ is the confidence threshold. $q_j$ is the pseudo-label of $x^{(u)}_j$, $q_j=argmax(f(\mathcal{A}_w(x^{(u)}_j);\theta))$. $\mathcal{A}_w(\cdot)$, $\mathcal{A}_s(\cdot)$ denote weak~\cite{xie2020unsupervised} and strong augmentation~\cite{cubuk2020randaugment, devries2017improved} for unlabeled data, respectively.

Examples Definitions:
\[
\left\{
\begin{array}{l}
\text{Easy examples: } \arg\max (\hat{p}(x_i)) \ge 0.95 \\
\text{Hard examples: } \tau \le \arg\max (\hat{p}(x_i)) < 0.95 \\
\text{Ultra-hard examples: } \arg\max (\hat{p}(x_i)) < \tau
\end{array}
\right.
\]
$\hat{p}(x_i)$ is the predicted probability for $x^{(u)}_i$ on the $i$-th class.

\subsection{Online Hard Example Mining and Learning} 
\label{sec:ohem}
In CISSL, it is common to set a high confidence threshold such as $\tau=0.95$, in this paper, we set $\tau=0.7$. The unlabeled pairs are used only when pseudo-labels' confidence exceeds the setting threshold. While using pseudo-labeled examples with high confidence for training can filter out a large amount of noisy data, this practice also ignores many potentially valuable hard samples. These hard examples include indistinguishable majority-class examples as well as rare minority-class examples. However, they are precisely the key to improving the model's generalizability. Intuitively, the solution is to lower the confidence threshold to access more hard examples. However, learning but not increasing the focus on low-confidence examples will fail to reduce the uncertainty of generating pseudo-labels, which will affect the robustness of the model. To this end, we propose an entropy-based online mining algorithm for hard examples. The algorithm leverages the characteristic that the entropy of the predicted label probability distribution is higher for hard examples, and assigns larger weights to these examples to improve the model's focus on hard samples. Re-weighting for unlabeled data can be expressed as:
\begin{equation}
    \label{eq:ohem_base}
    w(x^{(u)}_i)=\mathcal{J} (\hat{p}(x_i))
\end{equation}
\begin{equation}
    \label{eq:ohem}
    \mathcal{J} (\hat{p}(x_i))=\frac{ {\textstyle \sum_{i}^{K}\hat{p}(x_i){\log}\hat{p}(x_i)} }{{\log}K} \cdot s+\xi
\end{equation}
where $\frac{ {\textstyle \sum_{i}^{K}\hat{p}(x_i){\log}\hat{p}(x_i)} }{{\log}K}$ denotes the entropy of the predicted probability distribution. $logK$ is introduced to normalize the entropy, where $K$ is the size of the output label space. In addition, we introduce the scaling factors $s$ and $\xi$, which can scale $w(x^{(u)}_i)$ to the range of $(\xi,s+\xi)$, here $\xi=1-s$. The operation considers that easy examples with tiny entropy will be reweighted to a value approximating 0, leading the model to over-focus on hard examples with high entropy values and ignoring easy examples with low entropy values. This situation harms the training process. The scaling factor ensures that the easy examples can be given appropriate weights, making the model focus on both hard and easy examples, thus increasing the stability of the training process.
Based on $\mathcal{L}_U$, we obtain the loss by reweighting the hard examples:
\begin{equation}
    \mathcal{L}_U=\frac{1}{B_u}\sum_{j=1}^{B_u}  \mathcal{M}(x^{u}_j) w(x^{(u)}_j)\mathcal{H} (f(\mathcal{A}_s( x^{(u)}_j);\theta),q_j)
\end{equation}

In addition, although lowering the confidence threshold, there will still be some hard samples whose confidence is below the threshold, and we call them ultra-hard samples. These examples are also vital to improving model performance. Thanks to ~\cite{nassar2023protocon}, the model can be accelerated to learn the ultra-hard examples by introducing an embedding alignment constraint to strengthen the consistency between the weak and strong views:
\begin{equation}
    \label{eq:ea}
    \mathcal{L}_{EA}=\frac{1}{B_u}\sum_{j=1}^{B_u}  (1-\mathcal{M}(x^{(u)}_j)) \mathcal{H} (\hat{\delta } (e^w_j/5T_e),\hat{\delta }(e^s_j/T_e))
\end{equation}
where $\hat{\delta }$ is sharpened softmax. $e^w_j$ and $e^s_j$ denote the embedding of unlabeled data for weak and strong views, respectively. $T_e$ is the temperature parameter. $\mathcal{L}_{EA}$ assists $\mathcal{L}_U$ in accelerating the learning of ultra-hard examples, improving the certainty of the model's discrimination on them to generate more reliable pseudo-labels.

\subsection{Pseudo-Label Certainty Enhancement}
\label{sec:plr}
When online mining and learning hard examples, we lower the confidence threshold for generating pseudo-labels to mine more hard samples. This process boosted the weight of the hard examples and their confidence in the prediction. However, the pseudo-labels generated based on the low confidence threshold still reduce their certainty. Inspired by ~\cite{oh2022daso}, constructing a class-balanced memory bank and using a mixture of pseudo-labels and semantic labels can enhance label robustness. However, previous studies ~\cite{oh2022daso, peng2023bmb} ensured that the embeddings in the memory bank were class-balanced, neglecting the importance of utilizing embedding confidence information. Moreover, the higher the confidence of an embedding, the higher the certainty of the corresponding pseudo-label, and vice versa.

To this end, we design a balanced confidence decay memory bank ($\mathcal{B} $). $\mathcal{B} $ sets the storage space of size $\hat{B_i}$ for $i$-th class, which guarantees that the embeddings are class-balanced. Furthermore, it considers the pseudo-labels' certainty of the embeddings during updating. The updated embedding is pushed into the memory bank only when its confidence exceeds the lowest confidence of the embedding stored in the current class. It replaces the embedding with the lowest confidence. Notably, only the strong view embeddings are received in the memory bank. However, some easy examples with high confidence (e.g., the confidence is 0.999) exist in both majority and minority classes, which are hardly surpassed by other examples. This phenomenon led to these easy examples staying in the memory bank, and it prefer to select these examples when the class-balanced classifier randomly sampled from the memory bank, which reduces the diversity of learnable examples. To address this problem, we set a decay factor $\beta$. For every n steps, we apply a confidence decay to all the embeddings stored in the memory bank. This method ensures the diversity of the samples in the memory bank. More details of the algorithm are available in Appendix {\color{red}B}. The formula for the confidence decay is as follows:
\begin{equation}
    {\sigma}^t_{k,i} = {\sigma}^{t-1}_{k,i}\cdot \beta
\end{equation}
Where ${\sigma}^t_{k,i}$ denotes the confidence of the $i$-th embedding in the $k$-th class in the memory bank at moment $t$, and $t-1$ denotes the $n$-steps before moment $t$.

Distinct from ~\cite{oh2022daso}, we use more $q_j$ to mix pseudo-labels at the early stage of training. The reason is that using non-robust semantic pseudo-label mixing with a higher proportion at the early stage of training severely degrades the quality of pseudo-labels. Therefore, we adopt a dynamic pseudo-label mixing approach:
\begin{equation}
    \label{eq:q_prime}
    q^{\prime}_j=(1-\gamma\cdot w_k)\cdot q_j+(\gamma\cdot w_k)\cdot \hat{q_j}
\end{equation}
$\hat{q_j}$ is the semantic pseudo-label obtained by the similarity between query embedding and the embedded prototypes of each category, $\hat{q_j}=\delta(dist(e^s, C)/T_p)$, where $e^s$ denotes the query embedding. $T_p$ is the temperature parameter. $dist(\cdot,\cdot)$ denotes the Euclidean distance, and the smaller the distance, the higher the similarity is. $C$ is the set of category prototypes, $C=\{c_k\}^K_{k=1}$, where $c_k$ is the prototype of the $k$-th class, which is represented as $c_k=\frac{1}{\hat{M}} \sum_{i=1}^{\hat{M}} e_{k,i}$, $e_{k,i}$ denotes the $i$-th embedding of the $k$-th class in the memory bank, and $\hat{M}$ is the size of the memory bank. In Eqn.~(\ref{eq:q_prime}), $\gamma$ is the stabilization function, $\gamma=\frac{\alpha \cdot epoch_{current}}{epoch_{total}} $, $\alpha$ is a hyperparameter. $w_k$ is the estimated class weight based on the unlabeled data distribution. The more volume of data in the class, the higher the corresponding weight. Note that $\tilde{m_k}$ is the normalized class distribution of the current pseudo-labels, which is the accumulation of $q^{\prime}$ over previous iterations, e.g. $\tilde{m_1}+\tilde{m_2}+\cdots+\tilde{m_K}=1$.

The $\mathcal{L}_U$ after label refinement is:
\begin{equation}
    \mathcal{L}_U=\frac{1}{B_u}\sum_{j=1}^{B_u}  \mathcal{M}(x^{u}_j) w(x^{(u)}_j)\mathcal{H} (f(\mathcal{A}_s( x^{(u)}_j);\theta),q^{\prime}_j)
\end{equation}

\subsection{Decoupling Learning for Unbiased Prediction.} 
For CISSL tasks, reducing classifier bias towards majority classes is crucial. Inspired by decoupled representation learning in \cite{lee2021abc, peng2023bmb, wei2023towards} we use the Logit Align ~\cite{menon2020long} to reduce the model's prediction bias towards head classes in a balanced classifier. The balanced entropy loss and consistency loss friendly to tail categories are as follows:
\begin{equation}
    \mathcal{L}^b_S = \frac{1}{B_l}  {\textstyle \sum_{i=1}^{B_l}} \mathcal{H} (\tilde{f}(x^{(l)}_i;\theta ), y^{(l)}_i)
\end{equation}
\begin{equation}
\label{eq:l_ub}
    \mathcal{L}^b_U = \frac{1}{B_u}\sum_{j=1}^{B_u}  \tilde{\mathcal{M}}(x^{(u)}_j) \mathcal{H} (\tilde{f}(\mathcal{A}_s( x^{(u)}_j);\theta),q^{\prime}_j)
\end{equation}
where $\tilde{\mathcal{M}}(x^{(u)}_j):=\mathbbm{1}(max(\delta(\tilde{f}(x^{(u)}_j))) - T_{b}\cdot log\pi>\tau)$, $log\pi$ denotes the distribution of all categories in the labels of valid pseudo-labels and labeled data that are jointly updated and maintained in each step.

In the training phase, the base FixMatch algorithm is first used to warm up the standard branch and memory bank. After that, the losses of the standard branch and the balanced branch are merged to train the whole model together, and the total loss can be formulated as follows:
\begin{small}
\begin{equation}
    \mathcal{L}_{total} = \underbrace {\mathcal{L}_S+\mathcal{L}_U+ \mathcal{L}_{EA}  }_{standard\ branch} +\underbrace{\mathcal{L}^b_{S}+\mathcal{L}^b_{U}}_{balanced\ branch} 
\end{equation}
\end{small}
\section{Experiments}
\begin{table*}[!ht]
\centering
\small
\setlength{\tabcolsep}{0.0001mm}
\begin{adjustbox}{width=0.93\textwidth}
\begin{tabular}{lcccccccc}
\toprule 
 & \multicolumn{4}{c}{\textbf{CIFAR10-LT}}             & \multicolumn{4}{c}{\textbf{CIFAR100-LT}}              \\
& \multicolumn{2}{c}{$\gamma_l=\gamma_u=100$} &\multicolumn{2}{c}{$\gamma_l=\gamma_u=150$}  &\multicolumn{2}{c}{$\gamma_l=\gamma_u=10$} 
&\multicolumn{2}{c}{$\gamma_l=\gamma_u=20$}  \\
\cmidrule(lr){2-5} \cmidrule(lr){6-9} 
& $N_1=500$ \quad & $N_1=1500$ \quad \quad & $N_1=500$ & $N_1=1500$ \quad\quad& $N_1=50$ \quad& $N_1=150$ \quad\quad& $N_1=50$ \quad& $N_1=150$  \\
\textbf{Method} & $M_1=4000$ \quad & $M_1=3000$ \quad \quad& $M_1=4000$ \quad& $M_1=3000$ \quad\quad& $M_1=400$ \quad& $M_1=300$ \quad\quad& $M_1=400$ \quad& $M_1=300$  \\
\midrule 
\textbf{Supervised}   &$47.3${\scriptsize $\pm 0.95$}&$61.9${\scriptsize $\pm 0.41$}&$44.2${\scriptsize $\pm 0.33$}&$58.2${\scriptsize $\pm 0.29$}&$29.6${\scriptsize $\pm 0.57$}&$46.9${\scriptsize $\pm 0.22$}&$25.1${\scriptsize $\pm 1.14$}&$41.2${\scriptsize $\pm 0.15$} \\
\textbf{\quad  w/LA}         &$53.3${\scriptsize $\pm 0.44$}&$70.6${\scriptsize $\pm 0.21$}&$49.5${\scriptsize $\pm 0.40$}&$67.1${\scriptsize $\pm 0.78$}&$30.2${\scriptsize $\pm 0.44$}&$48.7${\scriptsize $\pm 0.89$}&$26.5${\scriptsize $\pm 1.31$}&$44.1${\scriptsize $\pm 0.42$} \\
\midrule
\textbf{FixMatch}     &$67.8${\scriptsize $\pm 1.13$}&$77.5${\scriptsize $\pm 1.32$}&$62.9${\scriptsize $\pm 0.36$}&$72.4${\scriptsize $\pm 1.03$}&$45.2${\scriptsize $\pm 0.55$}&$56.5${\scriptsize $\pm 0.06$}&$40.0${\scriptsize $\pm 0.96$}&$50.7${\scriptsize $\pm 0.25$} \\
\quad w/DARP &$74.5${\scriptsize $\pm 0.78$}&$77.8${\scriptsize $\pm 0.63$}&$67.2${\scriptsize $\pm 0.32$}&$73.6${\scriptsize $\pm 0.73$}&$49.4${\scriptsize $\pm 0.20$}&$58.1${\scriptsize $\pm 0.44$}&$43.4${\scriptsize $\pm 0.87$}&$52.2${\scriptsize $\pm 0.66$}  \\
\quad w/DASO &$76.0${\scriptsize $\pm 0.37$}&$79.1${\scriptsize $\pm 0.75$}&$70.1${\scriptsize $\pm 1.81$}&$75.1${\scriptsize $\pm 0.77$}&$49.8${\scriptsize $\pm 0.24$}&$59.2${\scriptsize $\pm 0.35$}&$43.6${\scriptsize $\pm 0.09$}&$52.9${\scriptsize $\pm 0.42$}  \\
\quad w/Crest+ &$76.3${\scriptsize $\pm 0.86$}&$78.1${\scriptsize $\pm 0.42$}&$67.5${\scriptsize $\pm 0.45$}&$73.7${\scriptsize $\pm 0.34$}&$44.5${\scriptsize $\pm 0.94$}&$57.4${\scriptsize $\pm 0.18$}&$40.1${\scriptsize $\pm 1.28$}&$52.1${\scriptsize $\pm 0.21$}  \\
\quad w/BEM &$75.8${\scriptsize $\pm 1.13$}&$80.3${\scriptsize $\pm 0.62$}&$69.7${\scriptsize $\pm 0.91$}&$75.7${\scriptsize $\pm 0.22$}&$50.4${\scriptsize $\pm 0.34$}&$59.0${\scriptsize $\pm 0.23$}&$44.1${\scriptsize $\pm 0.18$}&$54.3${\scriptsize $\pm 0.36$}  \\
\midrule
\textbf{FixMatch + LA}      &$75.3${\scriptsize $\pm 2.45$}&$82.0${\scriptsize $\pm 0.36$}&$67.0${\scriptsize $\pm 2.49$}&$78.0${\scriptsize $\pm 0.91$}&$47.3${\scriptsize $\pm 0.42$}&$58.6${\scriptsize $\pm 0.36$}&$41.4${\scriptsize $\pm 0.93$}&$53.4${\scriptsize $\pm 0.32$}\\
\quad w/DARP &$76.6${\scriptsize $\pm 0.92$}&$80.8${\scriptsize $\pm 0.62$}&$68.2${\scriptsize $\pm 0.94$}&$76.7${\scriptsize $\pm 1.13$}&$50.5${\scriptsize $\pm 0.78$}&$59.9${\scriptsize $\pm 0.32$}&$44.4${\scriptsize $\pm 0.65$}&$53.8${\scriptsize $\pm 0.43$}  \\
\quad w/DASO &$77.9${\scriptsize $\pm 0.88$}&$82.5${\scriptsize $\pm 0.08$}&$70.1${\scriptsize $\pm 1.68$}&$79.0${\scriptsize $\pm 2.23$}&$50.7${\scriptsize $\pm 0.51$}&$60.6${\scriptsize $\pm 0.71$}&$44.1${\scriptsize $\pm 0.61$}&$55.1${\scriptsize $\pm 0.72$}  \\
\quad w/Crest+ &$76.7${\scriptsize $\pm 1.13$}&$81.1${\scriptsize $\pm 0.57$}&$70.9${\scriptsize $\pm 1.18$}&$77.9${\scriptsize $\pm 0.71$}&$44.0${\scriptsize $\pm 0.21$}&$57.1${\scriptsize $\pm 0.55$}&$40.6${\scriptsize $\pm 0.55$}&$52.3${\scriptsize $\pm 0.20$}  \\
\quad w/BEM &$78.6${\scriptsize $\pm 0.97$}&$83.1${\scriptsize $\pm 0.13$}&$72.5${\scriptsize $\pm 1.13$}&$79.9${\scriptsize $\pm 1.02$}&$51.3${\scriptsize $\pm 0.26$}&$61.9${\scriptsize $\pm 0.57$}&$44.8${\scriptsize $\pm 0.21$}&$56.1${\scriptsize $\pm 0.54$}  \\
\midrule
\textbf{FixMatch + ABC}      &$78.9${\scriptsize $\pm 0.82$}&$83.8${\scriptsize $\pm 0.36$}&$66.5${\scriptsize $\pm 0.78$}&$80.1${\scriptsize $\pm 0.45$}&$47.5${\scriptsize $\pm 0.18$}&$59.1${\scriptsize $\pm 0.21$}&$41.6${\scriptsize $\pm 0.83$}&$53.7${\scriptsize $\pm 0.55$}\\
\quad w/DASO &$80.1${\scriptsize $\pm 1.16$}&$83.4${\scriptsize $\pm 0.31$}&$70.6${\scriptsize $\pm 0.80$}&$80.4${\scriptsize $\pm 0.56$}&$50.2${\scriptsize $\pm 0.62$}&$60.0${\scriptsize $\pm 0.32$}&$44.5${\scriptsize $\pm 0.25$}&$55.3${\scriptsize $\pm 0.53$}  \\
\midrule
FixMatch w/SeMi (Ours)     &$\textbf{80.3}${\scriptsize $\pm 0.41$}&$\textbf{84.3}${\scriptsize $\pm 0.20$}&$\textbf{73.3}${\scriptsize $\pm 0.80$}&$\textbf{80.5}${\scriptsize $\pm 0.61$}&$\textbf{51.9}${\scriptsize $\pm 0.43$}&$\textbf{63.2}${\scriptsize $\pm 0.26$}&$\textbf{45.3}${\scriptsize $\pm 0.48$}&$\textbf{56.6}${\scriptsize $\pm 0.51$}\\

\bottomrule 
\end{tabular}
\end{adjustbox}
\caption{In a consistent ($\gamma_l=\gamma_u$) setting on CIFAR10/100-LT, we compare the accuracy (\%) of previous CISSL methods with our approach. \textbf{Bold} indicates the best results.}
\label{tab:consistent}
\end{table*}

\begin{table*}[!ht]
\centering
\small
\setlength{\tabcolsep}{0.0001mm}
\begin{adjustbox}{width=0.93\textwidth}
\begin{tabular}{lcccccccc}
\toprule 
 & \multicolumn{4}{c}{\textbf{CIFAR10-LT($\gamma_l \ne \gamma_u$)}}             & \multicolumn{4}{c}{\textbf{STL10-LT($\gamma_u = N/A$)}}              \\
& \multicolumn{2}{c}{$\gamma_u=1(uniform)$} &\multicolumn{2}{c}{$\gamma_u=1/100(reversed)$}  &\multicolumn{2}{c}{$\gamma_l=10$} 
&\multicolumn{2}{c}{$\gamma_l=20$}  \\
\cmidrule(lr){2-5} \cmidrule(lr){6-9} 
& $N_1=500$ \quad & $N_1=1500$ \quad \quad & $N_1=500$ & $N_1=1500$ \quad\quad& $N_1=150$ \quad& $N_1=450$ \quad\quad& $N_1=150$ \quad& $N_1=450$  \\
\textbf{Method} & $M_1=4000$ \quad & $M_1=3000$ \quad \quad& $M_1=4000$ \quad& $M_1=3000$ \quad\quad& $M_1=100k$ \quad& $M_1=100k$ \quad\quad& $M_1=100k$ \quad& $M_1=100k$  \\
\midrule
FixMatch   &$73.0${\scriptsize $\pm 3.81$}&$81.5${\scriptsize $\pm 1.15$}&$62.5${\scriptsize $\pm 0.94$}&$71.8${\scriptsize $\pm 1.70$}&$56.1${\scriptsize $\pm 2.32$}&$72.4${\scriptsize $\pm 0.71$}&$47.6${\scriptsize $\pm 4.87$}&$64.0${\scriptsize $\pm 2.27$} \\
\quad  w/DARP         &$82.5${\scriptsize $\pm 0.75$}&$84.6${\scriptsize $\pm 0.34$}&$70.1${\scriptsize $\pm 0.22$}&$80.0${\scriptsize $\pm 0.93$}&$66.9${\scriptsize $\pm 1.66$}&$75.6${\scriptsize $\pm 0.45$}&$59.9${\scriptsize $\pm 2.17$}&$72.3${\scriptsize $\pm 0.60$} \\
\quad  w/CREST         &$83.2${\scriptsize $\pm 1.67$}&$87.1${\scriptsize $\pm 0.28$}&$70.7${\scriptsize $\pm 2.02$}&$80.8${\scriptsize $\pm 0.39$}&$61.7${\scriptsize $\pm 2.51$}&$71.6${\scriptsize $\pm 1.17$}&$57.1${\scriptsize $\pm 3.67$}&$68.6${\scriptsize $\pm 0.88$} \\
\quad  w/CREST+         &$82.2${\scriptsize $\pm 1.53$}&$86.4${\scriptsize $\pm 0.42$}&$62.9${\scriptsize $\pm 1.39$}&$72.9${\scriptsize $\pm 2.00$}&$61.2${\scriptsize $\pm 1.27$}&$71.5${\scriptsize $\pm 0.96$}&$56.0${\scriptsize $\pm 3.19$}&$68.5${\scriptsize $\pm 1.88$} \\
\quad  w/DASO         &$86.6${\scriptsize $\pm 0.84$}&$88.8${\scriptsize $\pm 0.59$}&$71.0${\scriptsize $\pm 0.95$}&$80.3${\scriptsize $\pm 0.65$}&$70.0${\scriptsize $\pm 1.19$}&$78.4${\scriptsize $\pm 0.80$}&$65.7${\scriptsize $\pm 1.78$}&$75.3${\scriptsize $\pm 0.44$} \\
\quad w/BEM &$86.8${\scriptsize $\pm 0.47$}&$89.1${\scriptsize $\pm 0.75$}&$70.0${\scriptsize $\pm 1.72$}&$79.1${\scriptsize $\pm 0.77$}&$68.3${\scriptsize $\pm 1.15$}&$81.2${\scriptsize $\pm 1.42$}&$61.6${\scriptsize $\pm 0.98$}&$76.0${\scriptsize $\pm 1.51$}  \\
\quad w/SEVAL &$\textbf{90.3}${\scriptsize $\pm 0.61$}&$90.6${\scriptsize $\pm 0.47$}&$79.2${\scriptsize $\pm 0.83$}&$82.9${\scriptsize $\pm 1.78$}&$70.6^*${\scriptsize $\pm 0.54$}&$79.7^*${\scriptsize $\pm 0.49$}&$67.4${\scriptsize $\pm 0.69$}&$75.7${\scriptsize $\pm 0.36$}  \\
\quad w/SeMi (Ours)     &$89.2${\scriptsize $\pm 0.38$}&$\textbf{92.2}${\scriptsize $\pm 0.62$}&$\textbf{83.5}${\scriptsize $\pm 1.85$}&$\textbf{88.9}${\scriptsize $\pm 0.91$}&$\textbf{75.3}${\scriptsize $\pm 0.74$}&$\textbf{81.5}${\scriptsize $\pm 0.55$}&$\textbf{73.7}${\scriptsize $\pm 1.39$}&$\textbf{78.6}${\scriptsize $\pm 0.90$}\\

\bottomrule 
\end{tabular}
\end{adjustbox}
\caption{In real-world scenarios, it is challenging to maintain consistent distributions between unlabeled and labeled data. On CIFAR10-LT and STL10-LT, with the setting ($\gamma_l \ne\gamma_u$), we compare the accuracy (\%) of previous CISSL methods with our approach. $*$ indicates that this data was missing in the original paper and is obtained from their experimental code. $N/A$ denotes distribution unknown.}
\label{tab:uniform_reversde}
\end{table*}

\begin{table}[!ht]
\centering
\small
\setlength{\tabcolsep}{0.0001mm}
\begin{adjustbox}{width=0.95\columnwidth}
\begin{tabular}{lcccc}
\toprule 
 & \multicolumn{4}{c}{\textbf{CIFAR100-LT($\gamma_l \ne \gamma_u$)}}                          \\
& \multicolumn{2}{c}{$\gamma_u=1(uniform)$} &\multicolumn{2}{c}{$\gamma_u=1/100(reversed)$}    \\
\cmidrule(lr){2-5}  
& $N_1=50$ \quad & $N_1=150$ \quad \quad & $N_1=50$ & $N_1=150$ \quad  \\
\textbf{Method} & $M_1=400$ \quad & $M_1=300$ \quad \quad& $M_1=400$ \quad& $M_1=300$  \\
\midrule
FixMatch   &$45.5${\scriptsize $\pm 0.71$}&$58.1${\scriptsize $\pm 0.72$}&$44.2${\scriptsize $\pm 0.43$}&$57.3${\scriptsize $\pm 0.19$} \\
\quad  w/DARP &$43.5${\scriptsize $\pm 0.95$}&$55.9${\scriptsize $\pm 0.32$}&$36.9${\scriptsize $\pm 0.48$}&$51.8${\scriptsize $\pm 0.92$} \\
\quad  w/CREST &$43.5${\scriptsize $\pm 0.30$}&$59.2${\scriptsize $\pm 0.25$}&$39.0${\scriptsize $\pm 1.11$}&$56.4${\scriptsize $\pm 0.62$} \\
\quad  w/CREST+ &$43.6${\scriptsize $\pm 1.60$}&$58.7${\scriptsize $\pm 0.16$}&$39.1${\scriptsize $\pm 0.77$}&$56.4${\scriptsize $\pm 0.78$} \\
\quad  w/DASO &$53.9${\scriptsize $\pm 0.66$}&$61.8${\scriptsize $\pm 0.98$}&$51.0${\scriptsize $\pm 0.19$}&$60.0${\scriptsize $\pm 0.31$} \\
\quad w/SeMi (Ours)     &$\textbf{54.3}${\scriptsize $\pm 0.58$}&$\textbf{64.7}${\scriptsize $\pm 0.77$}&$\textbf{51.5}${\scriptsize $\pm 0.26$}&$\textbf{62.5}${\scriptsize $\pm 0.43$}\\

\bottomrule 
\end{tabular}
\end{adjustbox}
\caption{On CIFAR100-LT, with the setting ($\gamma_l \ne\gamma_u$), we compare the accuracy (\%) of previous CISSL methods with our approach.}
\label{tab:c100_uni_rev}
\end{table}



\begin{table}[!ht]
\centering
\begin{minipage}{0.23\textwidth}
    \centering
    \small
    \setlength{\tabcolsep}{0.0001mm}
    \begin{adjustbox}{width=0.9\columnwidth}
    \begin{tabular}{lcc}
    \toprule 
    Method & $\quad 32\times32$ & $\quad 64\times64$\\
    \midrule
    FixMatch   &\quad $29.7$&\quad $42.3$\\
    \quad  w/DARP &\quad $30.5$&\quad $42.5$ \\
    \quad  w/DARP +cRT &\quad $39.7$&\quad $51.0$\\
    \quad  w/CREST &\quad $32.5$&\quad $44.7$\\
    \quad  w/CREST++LA &\quad $40.9$&\quad $55.9$\\
    \quad  w/CoSSL &\quad $43.7$&\quad $53.9$\\
    \quad  w/TRAS &\quad $46.2$&\quad $54.1$ \\
    \quad  w/BEM &\quad $53.3$&\quad $58.2$\\
    \quad w/SeMi (Ours)     &\quad $\textbf{56.1}$ &\quad $\textbf{61.4}$\\
    \bottomrule 
    \end{tabular}
    \end{adjustbox}
    \caption{On the naturally long-tailed ImageNet127 dataset, we compare the accuracy (\%) of previous CISSL methods with our approach.}
    \label{tab:img127}
\end{minipage}
\hspace{0.04\textwidth} 
\begin{minipage}{0.18\textwidth}
    \centering
    \small
    \setlength{\tabcolsep}{0.0001mm}
    \begin{adjustbox}{width=0.9\columnwidth}
    \begin{tabular}{lcc}
    \toprule 
    Method & \quad C10 &\quad S10 \\
    \midrule
    SeMi (Ours)     &\quad $\textbf{80.1}$  &\quad $\textbf{75.6}$\\
    \quad w/o $\mathcal{B} $     &\quad $78.7$ &\quad $74.4$\\
    \quad w/o OHEML     &\quad $77.4$ &\quad $72.5$\\
    \quad w/o EA     &\quad $79.7$ &\quad $74.7$\\
    \quad w/o PLCE    &\quad $79.5$  &\quad $73.2$\\
    \quad w/o BC    &\quad $76.4$  &\quad $70.3$\\
    \bottomrule 
    \end{tabular}
    \end{adjustbox}
    \caption{Ablation experiments on C10-LT and S10-LT.}
    \label{tab:ablation}
\end{minipage}
\end{table}

\subsection{Experimental setup}
\textbf{Datasets.} We exhaustively evaluate the proposed method on several class-imbalance datasets, including CIFAR-10-LT~\cite{krizhevsky2009learning}, CIFAR100-LT~\cite{krizhevsky2009learning}, STL10-LT~\cite{coates2011analysis}, and ImageNet-127~\cite{fan2022cossl}. To create imbalanced versions of the datasets, we set $N_k = N_1\cdot \gamma_l ^{-\frac{k-1}{K-1} }$ for labeled data and $M_k = M_1\cdot \gamma_u ^{-\frac{k-1}{K-1} }$ for unlabeled data. More details in the Appendix {\color{red}A}.

Following common practice ~\cite{oh2022daso, wei2023towards} for \textbf{CIFAR-10-LT}, we evaluate the proposed method using various volumes of labeled and unlabeled data, e.g., set $N_1$=500, $M_1$=4000 and $N_1=1500$, $M_1=3000$, and at different imbalance ratios, $\gamma_l=\gamma_u=100$, $\gamma_l=\gamma_u=150$ and $\gamma_l=100$, $\gamma_u=1/100$. For \textbf{CIFAR100-LT}, we set $N_1=50$, $M_1=400$ and $N_1=150$, $M_1=300$. Set imbalance ratios to $\gamma_l=\gamma_u=10$, $\gamma_l=\gamma_u=20$ and $\gamma_l=10$, $\gamma_u=1/10$. For \textbf{STL10-LT}, following ~\cite{oh2022daso}, set $N_1=150$, $M_1 \approx 100k$ and $N_1=450$, $M_1\approx100k$. In this dataset, we cannot access the distribution of unlabeled data. Therefore, we set $\gamma_l=\{10, 20\}$. For \textbf{ImageNet-127}, it is an inherently class-imbalanced dataset with imbalance ratio $\gamma\approx286$. Following ~\cite{fan2022cossl}, we randomly select 10\% of training samples as the labeled set. We keep the original test set. In addition, the images are downsampled to $32\times32$, $64\times64$ pixels.

\noindent\textbf{Implementation Details.} Following previous work ~\cite{oh2022daso, wei2023towards}, our method was implemented using Wide ResNet-28-2~\cite{zagoruyko2016wide} for the CIFAR10-LT, CIFAR100-LT, and STL10-LT datasets and ResNet-50~\cite{he2016deep} for the ImageNet-127 dataset. To demonstrate the efficacy of our approach, we conducted a comparative analysis with several existing CISSL algorithms, including DARP~\cite{kim2020distribution}, CReST~\cite{wei2021crest}, DASO~\cite{oh2022daso}, ABC~\cite{lee2021abc}, BEM~\cite{zheng2024bem} and SEVAL~\cite{li2024learning}. The performance of all methods was evaluated using top-1 accuracy on the test set, with the mean and standard deviation reported for each method in three independent runs. More implementation details can be found in Appendix {\color{red}C}.

\subsection{Results on CIFAR10/100-LT and STL10-LT}
We conducted experiments on datasets with different distributions. We considered common scenarios when $\gamma_l=\gamma_u$. In addition, we set up uniform($\gamma_u=1$) and reversed($\gamma_u=1/100$) scenarios with $\gamma_l \ne\gamma_u$, respectively, to simulate real-world scenarios.

\begin{figure}[h]
  \centering
  \includegraphics[width=0.9\linewidth]{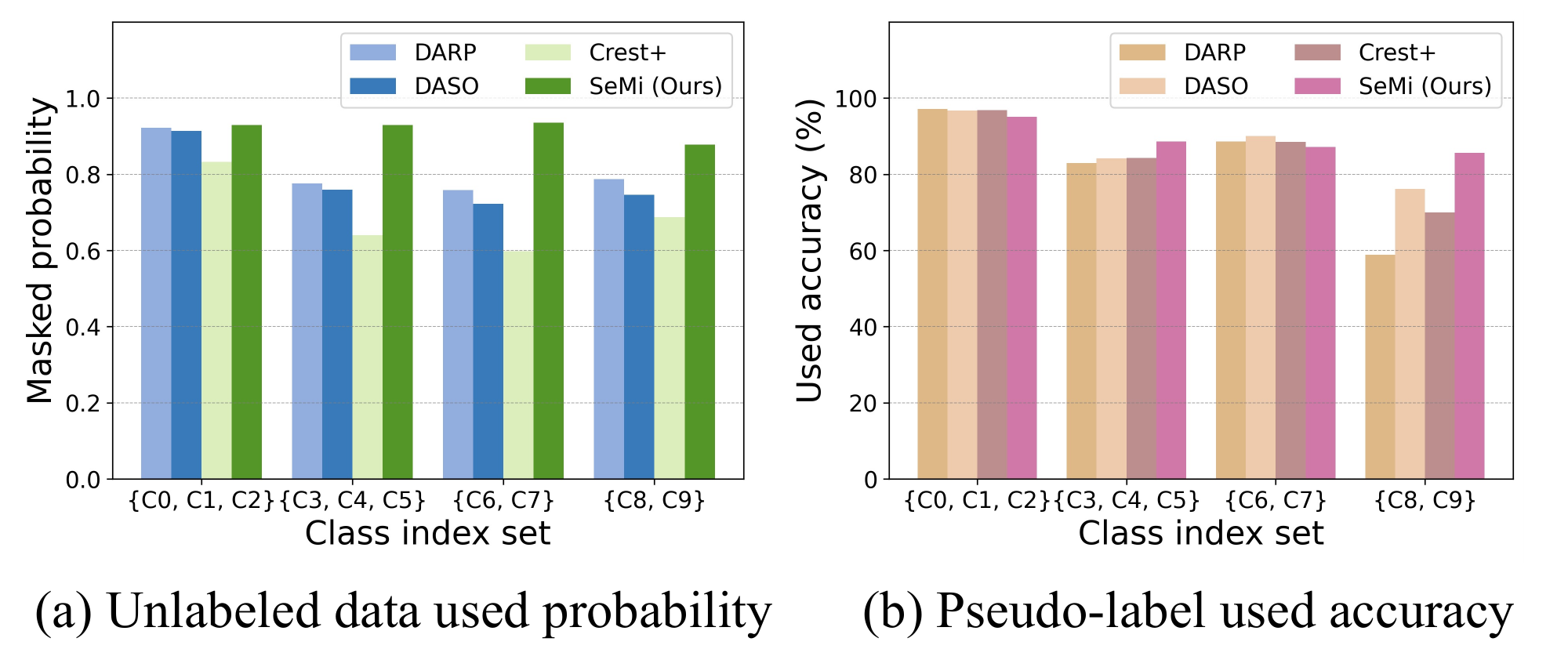}
  \caption{The CIFAR10-LT experiment with CISSL algorithms shows (a) average unlabeled data utilization across label sets and (b) pseudo-label accuracy aligning with ground truth. }
  \label{fig：mask_prob}
\end{figure}

\noindent\textbf{In case of} $\gamma_l=\gamma_u$. As shown in Table~\ref{tab:consistent}, we compare various state-of-the-art algorithms for addressing imbalance using Fixmatch as the baseline, including DARP, DASO, CReST+, BEM, and ABC. Additionally, we apply the Logits Align algorithm to these methods to balance their outputs. Our results achieve optimal performance across various settings, whether on CIFAR10-LT or CIFAR100-LT. Combined with the results in Fig.\ref{fig：mask_prob}, we enhance the utilization of unlabeled data from a new perspective by mining more hard examples. Our approach generates more accurate pseudo-labels through pseudo-label certainty enhancement, thereby improving the performance of tail classes.

\noindent\textbf{In case of} $\gamma_l\ne \gamma_u$. We conducted experiments using data that more closely resembles real-world scenarios. As shown in Table~\ref{tab:uniform_reversde} and Table~\ref{tab:c100_uni_rev}, our approach achieves state-of-the-art results across different settings on CIFAR10/100-LT and STL10-LT datasets. Notably, in challenging setups such as CIFAR10-LT ($\gamma_l \ne\gamma_u$, $\gamma_u=1/100$) and STL10-LT ($\gamma_u=N/A$, $\gamma_l=10$, $\gamma_l=20$, $N_1=150$), our method significantly outperforms others. Our algorithm handles these tasks in reverse scenarios where more samples in the unlabeled data can be mined to supplement tail samples in the labeled data. Additionally, in STL10-LT settings with limited labeled data, our algorithm can generate more accurate pseudo-labels from a small amount of labeled data, enhancing model generalization. More detailed analysis can be found in Appendix {\color{red}D}.

In the common scenario ($\gamma_l=\gamma_u$) with FixMatch as the baseline, our method shows an average improvement of 13.7\% on CIFAR10-LT and 12.9\% on CIFAR100-LT. In more realistic settings ($\gamma_l \ne\gamma_u$), it improves by 23.2\% on CIFAR10-LT, 14.07\% on CIFAR100-LT, and 31.1\% on STL10-LT. According to the above statistics, our method significantly improved in all scenarios compared to the baseline, especially on challenging datasets that closely resemble real-world scenarios.

\subsection{Results on ImageNet-127}
ImageNet127 is introduced in several studies~\cite{huh2016makes, wei2021crest, zheng2024bem, wei2023towards, wei2024transfer, fan2022cossl, oh2022daso, kim2020distribution}, dividing the 1000 classes of ImageNet into 127 based on the WordNet hierarchy. Unlike other datasets, ImageNet127 naturally exhibits a long-tailed distribution with an imbalance ratio of approximately 286 without artificial construction. Following previous methods~\cite{wei2023towards, fan2022cossl}, we resize the original images to $32\times$ or $64\times64$ pixels using Pillow's box method and randomly select 10\% of the training samples as the labeled set. The results shown in Table~\ref{tab:img127} indicate that SeMi achieves significant improvements on both image sizes, with absolute test accuracy gains of 88.9\% and 45.2\% compared to FixMatch. The results demonstrate SeMi's effectiveness on long-tailed test datasets.

\subsection{Comprehensive analysis of the method.}


\begin{figure*}[h]
  \centering
  \includegraphics[width=\textwidth]{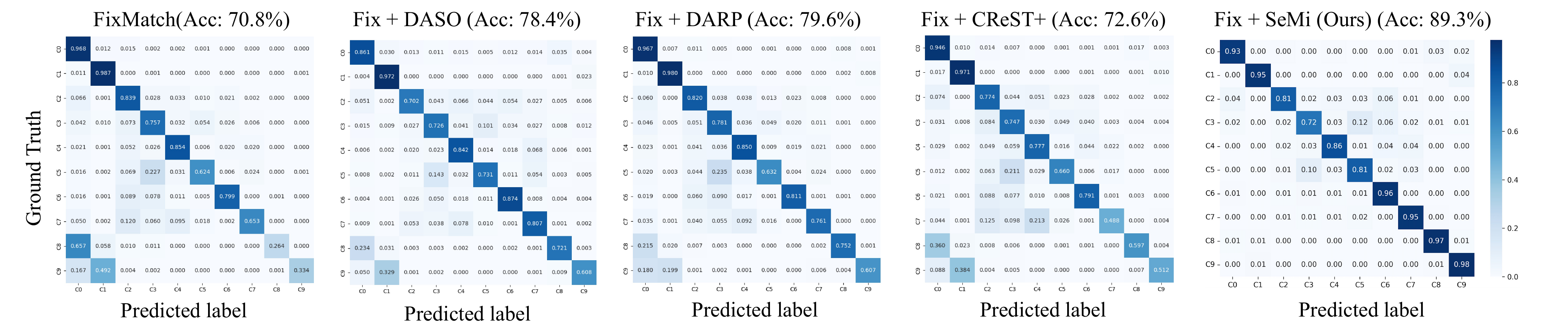}
  \caption{Confusion matrix comparison of multiple CISSL methods.}
  \label{fig:c10_dis}
\end{figure*}

\noindent\textbf{Ablation study on each component of SeMi.} 
As shown in Table~\ref{tab:ablation}, we conducted ablation experiments on SeMi using CIFAR10-LT with the setting ($\gamma_l = \gamma_u, \gamma_l=100, \gamma_u=100, N_1=500, M_1=4000$) and STL10-LT with the setting ($\gamma_l \ne \gamma_u, \gamma_l=10, \gamma_u=N/A, N_1=150, M_1=100k$). When we removed the Balanced Confidence Decay Memory Bank ($\mathcal{B} $), Online Hard Example Mining and Learning (OHEML) and Embeddings Align (EA) modules, the performance of SeMi dropped significantly. Additionally, when the Pseudo-Label Certainty Enhancement (PLCE) module was removed, the impact on CIFAR10-LT was less pronounced than on STL10-LT. Because STL10-LT has fewer labeled data, the quality of pseudo-labels generated from unlabeled data is more critical. Finally, we removed the Balanced Classifier (BC) module, which significantly impacted both datasets and indicated that it is critical for tail-biased output. The results demonstrate the effectiveness of SeMi.

\begin{figure}[h]
  \centering
  \includegraphics[width=0.94\linewidth]{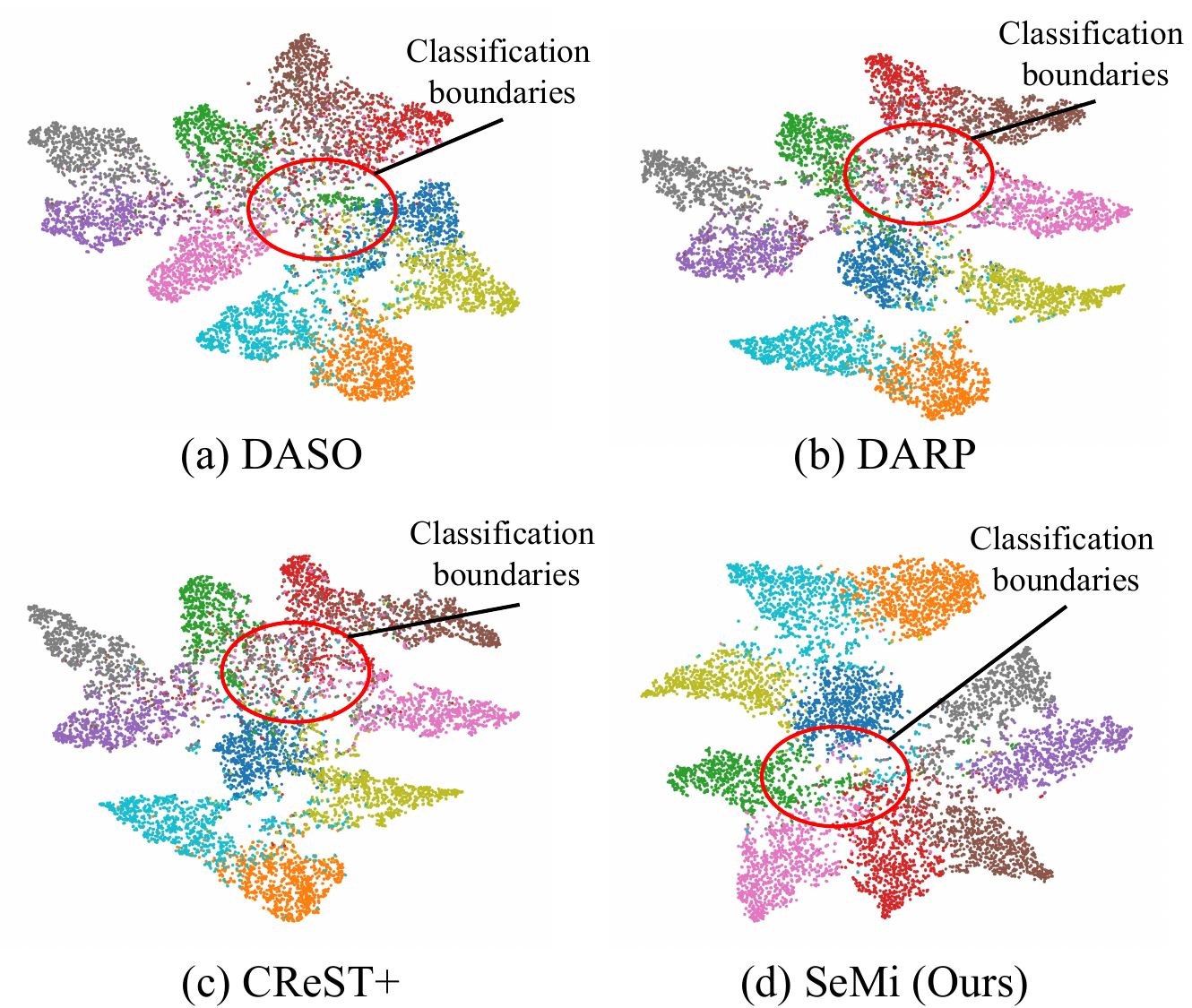}
  \caption{Comparison among various CISSL methods with t-SNE visualization.}
  \label{fig：features_vis}
\end{figure}

\noindent\textbf{Visualization of T-SNE.}
We use t-SNE ~\cite{van2008visualizing} to visualize representations on balanced test sets. We compare DASO, DARP, and CReST+ methods on CIFAR10-LT with the setting ($\gamma_l = \gamma_u, \gamma_l=100, \gamma_u=100, N_1=1500, M_1=3000$). The results in Figure~\ref{fig：features_vis} show that our methods generate clearer classification boundaries for the representations. Visualizations of the representations from these methods with the Logits Align algorithm are in Appendix {\color{red}E}.

\begin{figure}[h]
  \centering
  \includegraphics[width=0.94\linewidth]{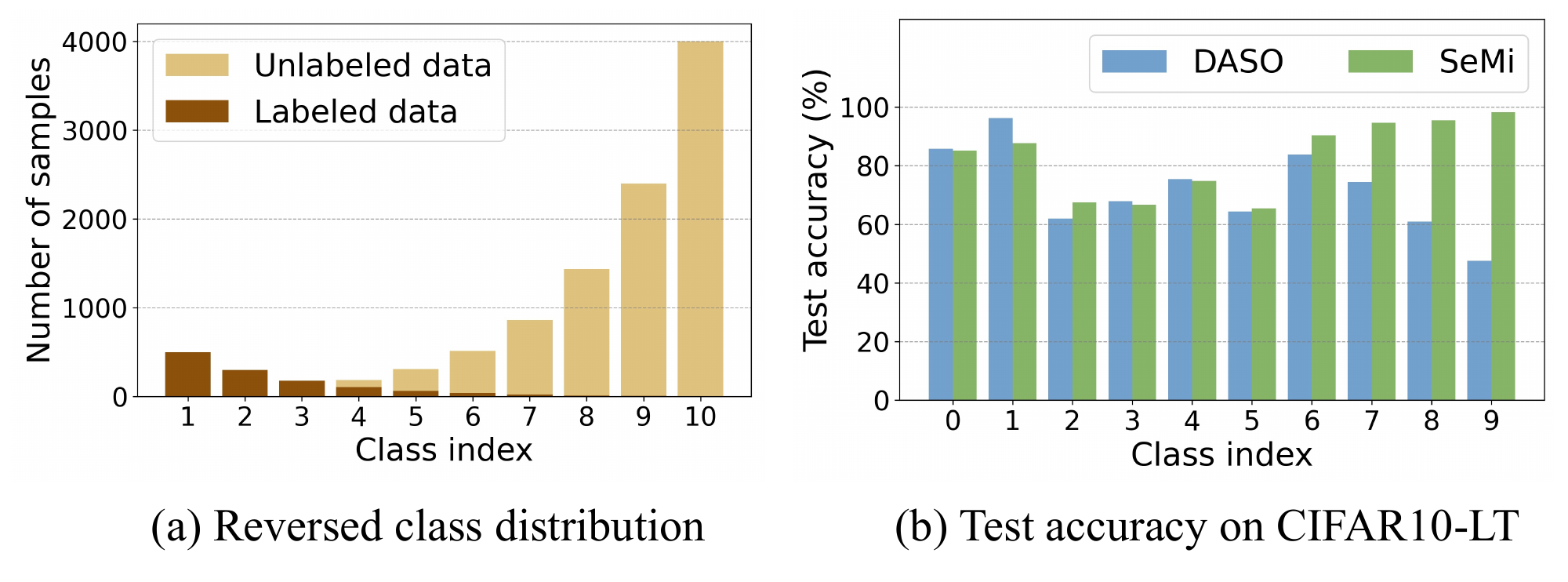}
  \caption{(a) Unlabeled and labeled data distribution. (b) Per-class accuracy: DASO vs. our method on a balanced test set.}
  \label{fig:semi_daso_acc}
\end{figure}

\noindent\textbf{Analysing results in reversed scenarios.} We conducted experiments on CIFAR10-LT under a challenging setting ($\gamma_l \ne \gamma_u$, with $\gamma_l=100$, $\gamma_u=1/100$, $N_1=500$, and $M_1=4000$), aiming to compare the performance of our SeMi algorithm with FixMatch, DASO, DARP, and CReST+. As shown in Figure~\ref{fig:semi_daso_acc}, our algorithm achieves a substantial improvement in tail-class performance by effectively utilizing the abundant tail-class samples from the unlabeled data. Further comparative analysis can be found in Appendix \textcolor{red}{F}.

\noindent\textbf{Confusion matrix analysis of predictions.} In another setting ($\gamma_l \ne \gamma_u$, with $\gamma_l=100$, $\gamma_u=1/100$, $N_1=1500$, and $M_1=3000$), analyzing the confusion matrices (Figure~\ref{fig:c10_dis}), SeMi exhibits significant advantages over the baseline methods, attaining an impressive accuracy of 89.3\%. In contrast, FixMatch lags with an accuracy of 70.8\%, showing considerable misclassifications, particularly in classes such as C5 and C8, where confusion is prominent. DASO and DARP incrementally improve to 78.4\% and 79.6\% accuracy, respectively, but still struggle with overlapping class boundaries and specific complex patterns, as evident in middle-range class errors. CReST+ achieves a slight accuracy gain to 72.6\%, yet falls short of DASO and DARP's precision. Notably, SeMi significantly outperforms these methods on tail classes (e.g., C7, C8, C9), demonstrating enhanced tail performance by effectively mining hard examples and refining pseudo-labels in CISSL.


\section{Conclusion}
Class-imbalanced semi-supervised learning poses a significant challenge in effectively utilizing imbalanced labeled and unlabeled data to improve classification performance for tail classes. This study introduces a novel approach based on hard example mining to address this issue. We improve the performance of tail categories by mining and leveraging more hard examples (e.g., the ones with in tail classes) in unlabeled data. Additionally, we improve the robustness and accuracy of pseudo-labels through pseudo-label with certainty enhancement, thereby boosting the overall model performance, even in scenarios with scarce tail-class labels. Extensive experiments performed on multiple benchmarks demonstrate the great advance on SOTA performance in CISSL obtained by the proposed approach.

\clearpage
{
    \small
    \bibliographystyle{ieeenat_fullname}
    \bibliography{main}
}


\end{document}